\documentclass[10pt,twocolumn,letterpaper]{article}
\usepackage[pagenumbers]{cvpr}
\usepackage{iftex}
\ifPDFTeX
  \usepackage[T1]{fontenc}
  \usepackage{textcomp}
\fi
\usepackage{amsmath,amssymb,graphicx,booktabs,xcolor}
\usepackage{placeins}
\usepackage{multicol}
\definecolor{cvprblue}{rgb}{0.21,0.49,0.74}
\usepackage[pagebackref,breaklinks,colorlinks,allcolors=cvprblue]{hyperref}
\def\confName{CVPR}
\def\confYear{2027}
\newcommand{\sg}{\operatorname{sg}}

\title{Dynamics Forcing: Mitigating Error Accumulation\\in Causal Video World Models}
\author{Xin Zhou$^{1}$ \quad Cong Miao$^{2}$\\
$^{1}$Astronex Robotics \quad $^{2}$Nanjing University of Information Science and Technology}

\begin{document}
\maketitle

\begin{abstract}
Causal video world models generate block by block from their own previous outputs. Small
errors in early blocks are fed back as history and grow until the scene and the robot fall apart.
Teacher forcing trains on clean ground-truth history, so the model never sees the history it will
actually condition on, and it also finds a shortcut: copying the previous frame instead of predicting
how the scene changes. We present Dynamics Forcing, a training scheme that keeps real video as the only
supervision and reshapes how history is presented. History is perturbed in three ways: random noising,
self-resampling by the model itself, and joint denoising, in which the most recent blocks are still
partly noisy when the next block is predicted. On top of this we move history guidance and first-frame
anchor guidance from inference into training, so that the model is pushed to use its history and its
reference frame rather than ignore them. Dynamics Forcing needs no teacher model, no distillation, and
no additional model at inference. Applied to a 5B video model with LoRA on 77 hours of manipulation
video on a single GPU, it better matches unseen real continuations and reduces error accumulation.
\end{abstract}

\section{Introduction}
A causal world model for robot manipulation must be rolled forward: given a first frame and
an instruction, it must produce the next seconds of video, then the seconds after that, each conditioned
on its own earlier output. Bidirectional video diffusion models generate a whole clip at once and cannot
be used this way. Recent work therefore converts them into causal, block-autoregressive
generators~\cite{chen2024diffusionforcing,yin2025causvid,huang2025selfforcing,yang2025longlive}.

Two problems appear as soon as such a model is rolled out.

\textbf{Error accumulation.} Under teacher forcing the history is always clean ground truth. At
inference it is the model's own output. The gap is small for the first block and compounds afterwards.
Our rollout analysis shows a consistent decline in semantic and patch similarity as generated history
is fed back over nine blocks (Sec.~\ref{sec:analysis}).

\textbf{Shortcut dynamics.} Adjacent frames of a manipulation video are nearly identical. A model that
copies its history already achieves a low denoising loss, so it has little pressure to learn what
actually changes. We measure this with the \emph{motion ratio}: the amount of motion in the generated
video divided by that of the real continuation. Values below one indicate insufficient motion, while
values above one indicate excessive motion, often caused by accumulated rollout errors.

Existing remedies either distil from a bidirectional teacher through distribution
matching~\cite{yin2025causvid,huang2025selfforcing}, which needs a second large model and inherits its
dynamics, or inject generated errors back into training. We ask a narrower question: \emph{how far can
one get with real video as the only supervision, by changing only how history is shown to the model?}

We address these issues with Dynamics Forcing, which has two parts.
(i) \emph{History perturbation.} The target block is always supervised by real video, but the history
it conditions on is perturbed by random noise, by the model's own short resampling of
it~\cite{guo2025resampling}, or by being denoised jointly with the target in a sliding window.
(ii) \emph{Training-time guidance.} History guidance~\cite{song2025historyguided} and a new first-frame
anchor guidance are applied to the training target instead of at inference, in the manner of model
guidance~\cite{tang2025modelguidance}. The model is trained to predict what guidance would have produced,
so that reliance on history and on the reference frame is learned rather than added afterwards.

Our contributions:
\begin{itemize}\setlength\itemsep{0pt}
\item A study of why causal video models trained with teacher forcing accumulate error and misestimate
motion over long rollouts.
\item Dynamics Forcing, a teacher-free training scheme that combines three history perturbations with a
joint-denoising noise rule that keeps older blocks cleaner than newer ones.
\item Training-time history guidance and first-frame anchor guidance, which reduce appearance drift of
the robot and objects over long rollouts.
\item Experiments on held-out long-horizon manipulation rollouts and VBench-I2V with a 5B model trained
on one GPU. On held-out manipulation videos, the method raises semantic similarity from 0.717 to 0.804
and patch similarity from 0.613 to 0.733 over teacher forcing. On VBench-I2V, it improves the starting
model's Total from 85.15 to 90.02.
\end{itemize}

\section{Related Work}
\textbf{Video world models for embodied AI.} Large video generators are increasingly used as simulators
for manipulation~\cite{cosmos,wan2025,simforcing,gigaworld1}. Cosmos~\cite{cosmos} provides a family of
world foundation models for physical AI; Wan~\cite{wan2025}, LTX-Video~\cite{ltxvideo},
CogVideoX~\cite{cogvideox}, and HunyuanVideo~\cite{hunyuanvideo} are open image-to-video generators that
serve as the priors from which such world models are typically post-trained. On the embodied side,
RBench/RoVid-X~\cite{rbench} is a large robotic video corpus with image--text evaluation pairs,
SimForcing~\cite{simforcing} distils simulation motion priors into real-domain robot world models, and
GigaWorld-1~\cite{gigaworld1} builds world models specifically for robot policy evaluation. UniSim
combines heterogeneous image, navigation, and manipulation data to learn an interactive real-world
simulator~\cite{yang2023unisim}, while IRASim conditions video prediction on robot trajectories to
capture fine-grained contacts~\cite{zhu2025irasim}. VideoWorld instead studies what task knowledge can
be acquired directly from unlabeled videos~\cite{ren2025videoworld}. Recent benchmarks show that even
strong video world models still exhibit non-physical deformation and unstable contact prediction in
robot manipulation~\cite{jiang2026robowmbench}. Most of these models are bidirectional and produce fixed
clips; world-model rollout requires causal generation, which is the setting we study.

\textbf{Autoregressive video diffusion.} Diffusion Forcing~\cite{chen2024diffusionforcing} assigns
independent noise levels to frames. CausVid~\cite{yin2025causvid} and Self
Forcing~\cite{huang2025selfforcing} distil a bidirectional teacher into a few-step causal student;
LongLive~\cite{yang2025longlive} extends this to long interactive generation with a frame sink and KV
re-caching. These teacher-distilled approaches require a second large model during training and inherit
its learned dynamics. Rolling Forcing~\cite{rollingforcing} instead combines rolling-window joint
denoising with an attention sink for multi-minute streaming. Our method uses real video targets without
a teacher or distribution-matching loss and is complementary to inference-time streaming techniques.
Beyond robot video, GameNGen~\cite{valevski2024gamengen},
DIAMOND~\cite{alonso2024diamond}, and Genie~\cite{bruce2024genie} demonstrate that autoregressive or
action-conditioned generative models can serve as interactive environments. Their long-horizon
stability likewise depends on keeping generated histories within the model's training distribution.

\textbf{Exposure bias and history corruption.} The mismatch between ground-truth context at training
time and generated context at inference is the classical exposure-bias problem. Scheduled
sampling~\cite{bengio2015scheduled} gradually replaces ground-truth tokens with model samples, whereas
Professor Forcing~\cite{lamb2016professor} aligns the free-running and teacher-forced dynamics. In
diffusion video models, noise-augmenting the context provides a continuous analogue.
Resampling Forcing~\cite{guo2025resampling} lets the model resample its own history during training, so
the history carries the model's own errors while the target stays real. We adopt it as one of three
perturbations and show that it is needed for correct motion, but not sufficient on its own: resampling
alone still lets the model copy its recent, only mildly corrupted history, which is why we add random
noising and a joint-denoising window that keeps older blocks cleaner than newer ones.

\textbf{Guidance.} Classifier-free guidance~\cite{ho2022cfg} and history
guidance~\cite{song2025historyguided} amplify a condition at inference at the cost of an extra forward
pass. Model guidance~\cite{tang2025modelguidance} folds the guidance term into the training target. We
apply this idea to history and to the reference frame of a causal video model, so that the guided
prediction is learned during training and no history-guidance pass is added at inference.

\section{Method}
\begin{figure*}[t]\centering
\includegraphics[width=\linewidth]{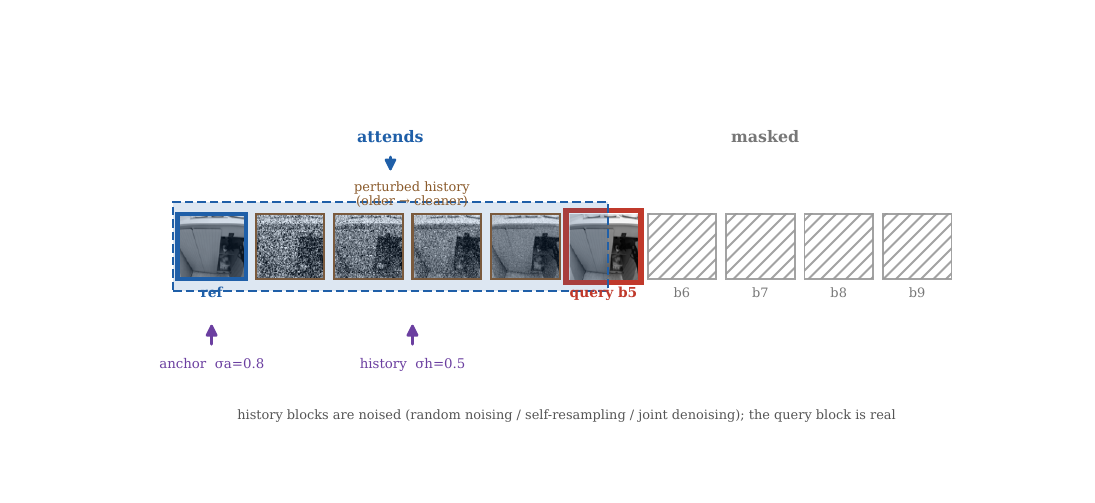}
\caption{Overview of Dynamics Forcing. Top: the block-causal attention pattern over latent blocks, where
each block attends to the reference frame, all earlier blocks and itself, and future blocks are masked.
Middle: the three history perturbations (random noising, self-resampling, joint denoising) applied to
the history blocks while the target block stays real. Bottom: the training-time history guidance
($\sigma_h{=}0.5$) and first-frame anchor guidance ($\sigma_a{=}0.8$), which add no forward pass beyond
the model's standard classifier-free-guidance inference.}\label{fig:pipeline}
\end{figure*}

Figure~\ref{fig:pipeline} summarizes the complete training construction. The top row defines the
block-causal information flow used at both training and inference. The middle row shows how the three
history perturbations replace clean teacher-forced context with imperfect rollout-like context while
leaving the supervised target real. The bottom row shows the two training-time guidance branches that
strengthen dependence on history and on the first-frame appearance anchor without adding a separate
history-guidance pass at inference.

\subsection{Preliminaries}
We start from a pretrained text-and-image-to-video flow-matching model~\cite{lipman2023flowmatching} and make its self-attention
block-causal. A clip is encoded into $F{=}72$ latent frames, grouped into $B{=}9$ blocks of 8; frame 0
is the clean reference image. Block $b$ attends to the reference frame, to all earlier blocks and to
itself. With $x^b_t=(1-t)\,x^b_0+t\,\epsilon$ and velocity target $v^b=\epsilon-x^b_0$, teacher forcing
minimises
\begin{equation}
\mathcal{L}_{\mathrm{TF}}=\mathbb{E}\,\big\|f_\theta(x^b_t,t\mid h^{<b},c)-v^b\big\|^2 ,
\end{equation}
where $h^{<b}=x^{<b}_0$ is clean history and $c$ the instruction. At inference blocks are generated one
after another with a KV cache.

\subsection{History perturbation}
The target is always real video. Only $h^{<b}$ changes. Let
$\widetilde h^{<b}=P(h^{<b};\xi)$ denote the perturbed history produced with random state $\xi$. Our
base objective is
\begin{equation}
\mathcal{L}_{\mathrm{DF}}=\mathbb{E}\,\big\|f_\theta(x^b_t,t\mid
\widetilde h^{<b},c)-v^b\big\|^2 .
\label{eq:df}
\end{equation}
No loss is applied to the history blocks, and all model-generated history is detached before being used
as context. Consequently, the supervision remains ordinary flow matching on real targets; the method
changes the conditioning distribution rather than introducing a teacher or a discriminator. In our
stochastic pipeline, random noising defines the base history corruption, while self-resampling and the
joint-denoising window are activated with the probabilities below.

\textbf{Random noised history.} Each history block is noised to a level $s$ and the model is told $s$
through that block's timestep. The level is tied to the target's level: when the target is at high noise
($t>0.6$) history is nearly clean, $s\in[0,0.1]$; when the target is at low noise the history is noisier,
$s\in[0.3,0.6]$. This schedule preserves informative context while preventing the network from treating
recent history as an exact copy target.

\textbf{Self-resampled history.} Following~\cite{guo2025resampling}, with probability 0.7 the history
is partially noised and denoised again by the current model in three steps, without gradient. The
history then carries the model's own structured errors while the target stays real. Detaching this
branch prevents the network from reducing Eq.~\eqref{eq:df} by changing how the context itself is
generated; it must instead learn to predict correctly from the imperfect context it receives.

\textbf{Joint denoising.} With probability 0.5 the last $W{=}3$ blocks form a window that is denoised
together. Noise levels decrease towards the past: if the newest block is at $t_k$, the block before it
is at
\begin{equation}
t_{k-1}=\operatorname{clip}\!\big(t_k-\delta,\;0.15,\;0.45\big),\qquad \delta\sim U(0.15,0.4).
\end{equation}
Window blocks enter the history in their noisy state. The lower bound keeps recent history from being
clean enough to copy; the upper bound keeps it informative. We found both bounds necessary: an upper
bound of 0.6 made the robot arm fade. Blocks outside the window remain available as older context, so
the model retains long-range appearance information while learning to bridge uncertainty at the current
generation boundary.

\subsection{Training-time history guidance}
History guidance contrasts a prediction made with full history against one made with weakened history.
We compute, without gradient, a weak-history prediction $f^{\mathrm{weak}}_\theta$ in which every
history block is re-noised to $\sigma_h{=}0.5$, and train
\begin{equation}
\mathcal{L}_{\mathrm{HG}}=\mathbb{E}\,\big\|f_\theta-w\,\sg\!\big(f_\theta-f^{\mathrm{weak}}_\theta\big)-v\big\|^2 ,
\label{eq:hg}
\end{equation}
with $w{=}1$, applied to half of the samples. The network output therefore converges to the guided
prediction, and no second pass is spent on history guidance at inference. The stop-gradient is
important: the weak branch defines a target direction but receives no gradient from the guidance term,
which avoids a degenerate solution in which both branches move together and the contrast disappears.

\subsection{First-frame anchor guidance}
Over long rollouts the robot and objects drift in appearance: in our runs a black arm gradually turned
white. The reference frame is the only place where the true appearance is given, so we extend the weak
branch of Eq.~\eqref{eq:hg}: the reference frame is also re-noised, to $\sigma_a{=}0.8$. The guidance
term then also amplifies what the clean first frame contributes.
At inference the same contrast is folded into standard classifier-free guidance: its unconditional
branch sees the reference frame re-noised to $\sigma_a$, so the anchor adds no forward pass. The
anchor is deliberately limited to the reference frame: applying the same strong condition to every
history block would favor appearance preservation at the expense of motion, whereas a single fixed
anchor separates identity information from the evolving state carried by recent blocks.

\subsection{Inference}
Blocks are generated autoregressively with KV caching, 8 sampling steps, shift 5, and classifier-free
guidance scale 5 at $832{\times}480$, 24\,fps.

\subsection{Why the three perturbations are complementary}
Each perturbation targets a different failure of teacher forcing, which is why we use all three rather
than treating them as alternatives. Random noising makes clean ground-truth history unavailable as an
exact copy source, so the network must predict change instead of repeating the previous block.
Self-resampling replaces noised history with history the model itself has denoised, so the errors it
feeds back are the errors it will make at inference, not isotropic noise. Joint denoising couples the
most recent blocks to the target through their shared noise schedule. The supplementary material
summarizes the distinct failure modes observed when each component is removed.

\section{Experiments}
\subsection{Setup}
\textbf{Model.} We start from Astronex-World~\cite{astronexworld}, a block-causal world model built on
Wan2.2-TI2V-5B~\cite{wan2025}, and train LoRA~\cite{hu2022lora}
adapters of rank 64. We use AdamW~\cite{loshchilov2019adamw} with learning rate $10^{-4}$, batch size 1, and gradient clipping at
1.0 on one 48\,GB GPU. The supplementary material gives the remaining implementation details.

\textbf{Data.} We use only publicly released datasets. The training set contains 21{,}124 clips
(18{,}012 unique, about 77 hours) of robot manipulation and egocentric human manipulation video,
including RoVid-X; 26\% are human videos, and hard-task clips account for 52\% after oversampling. The
held-out evaluation trajectories and tasks are excluded from training.

\textbf{Benchmarks and metrics.} For held-out manipulation videos we report DINOv2 semantic and patch
similarity, PSNR, and motion ratio. We additionally evaluate general image-to-video quality on
VBench-I2V~\cite{huang2024vbench,huang2025vbenchpp}.

\textbf{Held-out rollouts.} We evaluate zero-shot on trajectories and tasks that are never seen in
training. Each model receives only the first frame and the instruction and generates the continuation,
which is compared with the real video of the same clip. The real continuation is used only for
evaluation, never as model input or fine-tuning supervision.

\textbf{Scoring model.} Generated and real videos are scored with a frozen DINOv2 ViT-B/14
encoder~\cite{oquab2024dinov2}; no component of the scorer is trained on our data. Frames are sampled at the
same time stamps from both videos and resized to $224{\times}392$.
\emph{Semantic similarity} is the cosine similarity between the class tokens of corresponding frames,
averaged over time with a linear weight that doubles from the first to the last frame, so that late
frames count more.
\emph{Patch similarity} matches every patch token of a generated frame to its most similar patch in the
real frames within $\pm2$ sampled frames and averages the worst 20\% of matches; it penalises local
artefacts such as a deformed gripper that the class token does not see.
\emph{Motion ratio} is the mean absolute frame difference of the generated video divided by that of the
real video, both downsampled to $60{\times}104$.
We also report PSNR.

\subsection{VBench-I2V comparison}
\label{sec:vbench}
We also compare the model on VBench-I2V~\cite{huang2024vbench,huang2025vbenchpp} against Astronex-World~\cite{astronexworld}
and open image-to-video models, scored with the official code on the official prompt order, without
sampling or selection. Camera motion is excluded from the I2V Score and the Total (official
normalization and weighting); peer scores are as publicly reported under their respective official
configurations. Table~\ref{tab:vbench} reports the quantitative comparison. Ours posts the highest
Total and Quality and leads on I2V background and imaging quality, while dynamic degree remains the
weakest axis. The comparison includes bidirectional full-clip models as well as the causal starting
model and Dynamics Forcing.

\begin{table*}[t]\centering\footnotesize
\resizebox{\textwidth}{!}{%
\begin{tabular}{llccccccccccc}\toprule
Model & Temporal attention & I2V Subj. & I2V BG & Subj. Cons. & BG Cons. & Smooth. & Dyn. & Aesth. & Imag. & Quality & I2V Score & Total \\ \midrule
Astronex-World~\cite{astronexworld} & Causal & 91.08 & 97.99 & 88.51 & 96.09 & 96.74 & 51.22 & 57.69 & 68.65 & 76.88 & 93.41 & 85.15 \\
LTX-Video 0.9.7~\cite{ltxvideo} & Bidirectional & 98.97 & 99.20 & 97.87 & 98.39 & 99.52 & 24.80 & 64.59 & 70.24 & 80.32 & 98.86 & 89.59 \\
Wan2.2-TI2V-5B~\cite{wan2025} & Bidirectional & 97.79 & 98.92 & 94.31 & 96.14 & 96.60 & 58.53 & 61.95 & 71.24 & 79.96 & 97.97 & 88.97 \\
Cosmos-Predict2-14B~\cite{cosmos} & Bidirectional & 97.93 & 99.02 & 94.01 & 95.83 & 98.50 & 42.60 & 64.15 & 70.40 & 79.80 & 98.12 & 88.96 \\
CogVideoX1.5-5B-I2V~\cite{cogvideox} & Bidirectional & 97.78 & 98.73 & 95.04 & 96.52 & 98.47 & 37.48 & 62.68 & 70.99 & 79.54 & 97.84 & 88.69 \\
HunyuanVideo-I2V~\cite{hunyuanvideo} & Bidirectional & 98.90 & 99.38 & 95.67 & 96.39 & 99.21 & 21.14 & 61.55 & 70.37 & 78.31 & 98.94 & 88.62 \\
Wan2.1-I2V-14B-720P~\cite{wan2025} & Bidirectional & 96.95 & 96.44 & 94.86 & 97.07 & 97.90 & 51.38 & 64.75 & 70.44 & 80.83 & 95.80 & 88.32 \\
\textbf{Ours (Dynamics Forcing)} & \textbf{Causal} & \textbf{98.90} & \textbf{99.52} & \textbf{95.48} & \textbf{97.32} & \textbf{99.13} & \textbf{44.31} & \textbf{63.06} & \textbf{71.40} & \textbf{81.01} & \textbf{99.04} & \textbf{90.02} \\
\bottomrule
\end{tabular}
}
\caption{VBench-I2V scores with camera motion excluded from the I2V Score and Total.
``Astronex-World''~\cite{astronexworld} is the starting model; Ours (Dynamics Forcing) is shown in the
last row. We evaluate both with the official code and use the publicly reported scores for the remaining
models. Temporal attention refers to the denoising transformer, not the video VAE.}\label{tab:vbench}
\end{table*}

\begin{figure*}[t]\centering
\includegraphics[width=\linewidth]{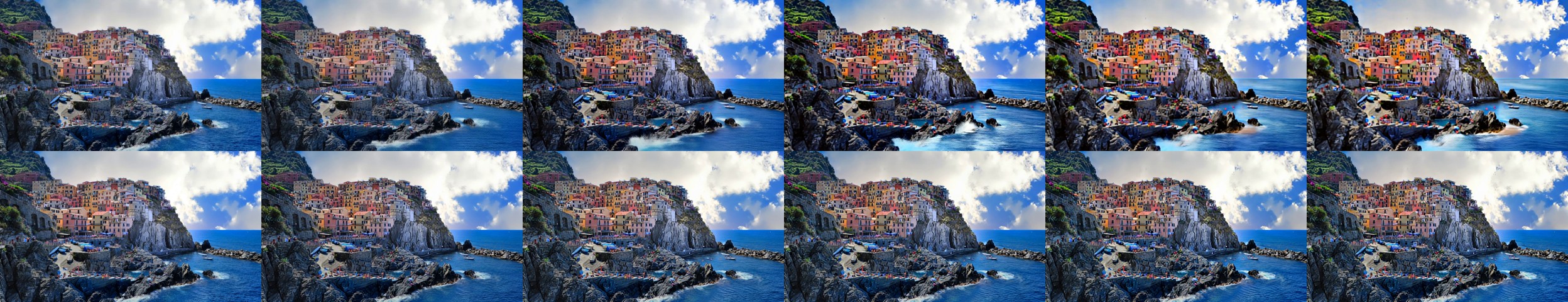}
\vspace{2pt}
\includegraphics[width=\linewidth]{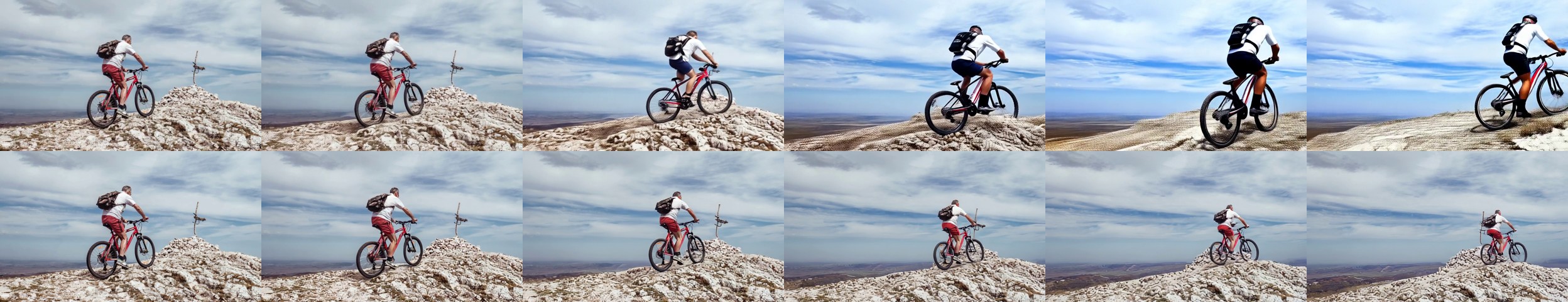}
\vspace{2pt}
\includegraphics[width=\linewidth]{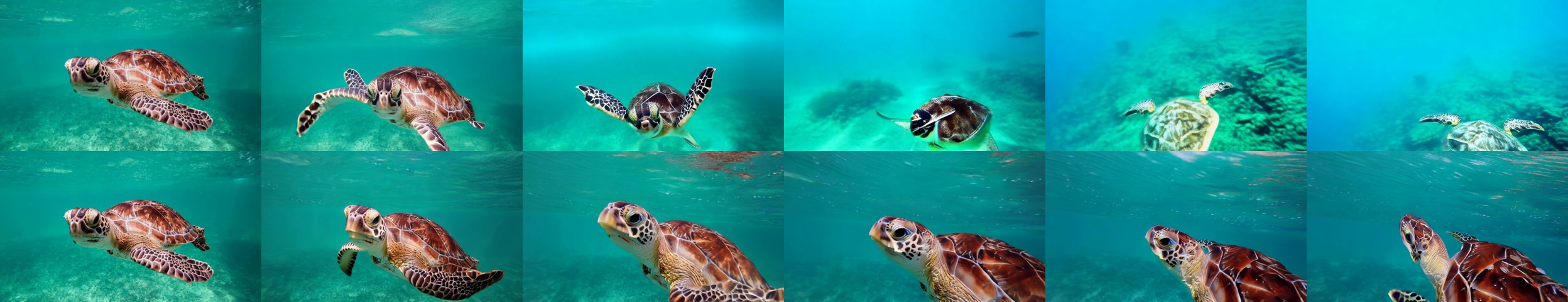}
\caption{VBench-I2V qualitative comparisons on three official prompts featuring seaside cliffs,
mountain biking, and a sea turtle (top to bottom). Within each example, the upper row is
Astronex-World and the lower row is Dynamics Forcing. Frames are sampled uniformly from each generated
video.}\label{fig:vbench-qual}
\end{figure*}

Figure~\ref{fig:vbench-qual} shows three representative prompts. The seaside scene remains spatially
stable in both models. In the mountain-bike and sea-turtle examples, the starting model changes the
rider, bicycle, or turtle appearance across time, whereas Dynamics Forcing retains the first-frame
identity and scene layout. The result matches the subject- and background-consistency gains in
Table~\ref{tab:vbench}.

Relative to the starting model, I2V subject consistency rises from 91.08 to 98.90 and I2V background
consistency from 97.99 to 99.52. These changes agree with the intended role of history and anchor
guidance: preserving the subject and scene specified by the reference frame during rollout. Quality
and imaging quality also increase, from 76.88 to 81.01 and from 68.65 to 71.40. The exception is
Dynamic Degree, which decreases from 51.22 to 44.31. Dynamic Degree measures visible motion magnitude,
while VBench-I2V has no paired real continuation for measuring whether that motion is correct. It
therefore assigns motion caused by rollout error and motion supported by the scene to the same axis.

Our paired held-out experiments make this distinction. On 50 unseen manipulation clips, Dynamics
Forcing has a motion ratio of 1.00 relative to the real continuation, whereas teacher forcing produces
$1.70\times$ as much motion. Over the 12-second evaluation, its ratios are 1.14,
1.04, and 0.80 across the three rollout segments, compared with 2.56, 1.43, and 0.87 for teacher
forcing; thus the model matches the real motion amount closely
for the first six blocks. Motion direction also improves: when both models are given real history, the
cosine between predicted and real pixel change rises from 0.402 to 0.482 over 50 clips and is higher on
41 of them (Sec.~\ref{sec:analysis}). The paired motion magnitude and direction results rule out scene
freezing as the source of the similarity gain. The lower VBench Dynamic Degree instead reflects the
reduction of excess or unsupported motion.

\subsection{Error accumulation}\label{sec:analysis}
We compare against the model our method starts from: the same block-causal network trained with plain
teacher forcing. Both models roll out nine blocks (12 seconds) on 16 held-out clips from the same first
frame, instruction and noise, and each block is scored against the real continuation
(Table~\ref{tab:acc}). Dynamics Forcing stays closer to the real video in every third of the rollout on
all three measures. The gain is largest in the first blocks and narrows later: in blocks 7--9 patch
similarity is $0.595$ against $0.537$ and PSNR $16.3$ against $15.5$\,dB, while semantic similarity is
$0.645$ against $0.633$. Dynamics Forcing maintains a lower error level throughout the rollout.

\begin{figure*}[t]\centering
\includegraphics[width=\linewidth]{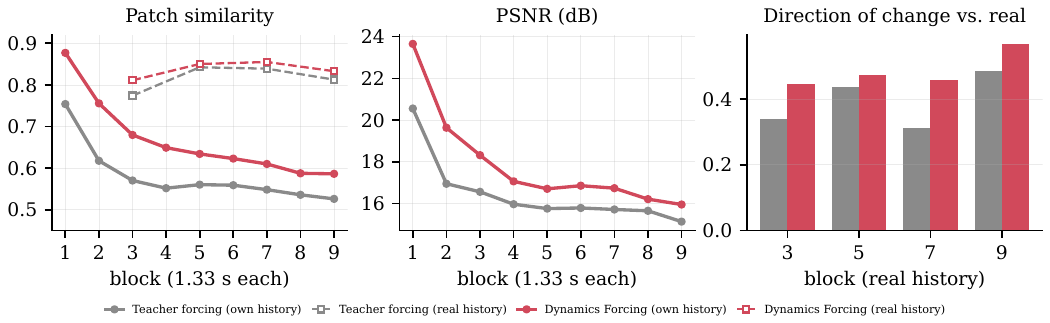}
\caption{Left, middle: similarity to the real video per block over a 12-second rollout (solid), and for
blocks generated from real history (dashed). Right: cosine between the predicted and the real change of
a block generated from real history. 16 held-out clips.}\label{fig:curves}
\end{figure*}

To separate accumulated error from the ordinary uncertainty of the future, we also generate single blocks
from \emph{real} history (blocks 3, 5, 7 and 9 of the same clips) and compare them with the same blocks
of the free rollout. The drop from real history to own history is the cost of accumulated error
(Fig.~\ref{fig:curves}, left). For patch similarity this drop is $0.210$ for Dynamics Forcing against
$0.266$ for teacher forcing (smaller in 46 of 64 block pairs); for semantic similarity the two are equal
($0.224$ against $0.226$). Given real history, the predicted change also points closer to the real one:
the cosine between predicted and real pixel change is $0.486$ against $0.393$ (higher in 52 of 64 pairs;
Fig.~\ref{fig:curves}, right). On all 50 held-out clips (block 5) the cosine is $0.482$ against $0.402$,
higher in 41 of 50 clips.
\begin{table}[!t]\centering\footnotesize
\resizebox{\columnwidth}{!}{%
\begin{tabular}{llccc}\toprule
Measure & Model & Blocks 1--3 & Blocks 4--6 & Blocks 7--9 \\ \midrule
Semantic$\uparrow$ & Teacher forcing & 0.786 & 0.681 & 0.633 \\
 & Dynamics Forcing & \textbf{0.842} & \textbf{0.715} & \textbf{0.645} \\ \midrule
Patch$\uparrow$ & Teacher forcing & 0.647 & 0.557 & 0.537 \\
 & Dynamics Forcing & \textbf{0.771} & \textbf{0.635} & \textbf{0.595} \\ \midrule
PSNR$\uparrow$ & Teacher forcing & 18.0 & 15.8 & 15.5 \\
 & Dynamics Forcing & \textbf{20.5} & \textbf{16.9} & \textbf{16.3} \\ \midrule
Motion ratio & Teacher forcing & 2.56 & 1.43 & 0.87 \\
 & Dynamics Forcing & 1.14 & 1.04 & 0.80 \\ \bottomrule
\end{tabular}
}
\caption{Similarity to the real video over a 12-second rollout (16 held-out clips, DINOv2 scorer).
Motion ratio: 1 = same amount of motion as the real video.}\label{tab:acc}
\end{table}

\subsection{Causal dynamics}
We evaluate causal dynamics as the agreement between the change predicted from a given history and the
change observed in the real continuation. We measure this agreement in two ways. First, from real
history: Sec.~\ref{sec:analysis} shows that the direction of the predicted change
agrees better with the real change than for teacher forcing (Fig.~\ref{fig:curves}, right). Second, from
the first frame alone: every evaluation trajectory and task is held out, so similarity to the real
continuation measures transfer to unseen interactions.
Table~\ref{tab:dyn} reports semantic and patch similarity to the real video
together with the motion ratio. Dynamics Forcing is closest to the real video on all three: semantic similarity
rises from $0.717$ to $0.804$ and patch similarity from $0.613$ to $0.733$ over teacher forcing, and the
amount of motion matches the real video (ratio $1.00$) where teacher forcing moves $1.7$ times too much.
In the fifth second the semantic similarity is $0.732$ against $0.661$. Self Forcing is evaluated
zero-shot with its released text-to-video weights as an off-the-shelf causal model reference.
\begin{table}[!t]\centering\footnotesize
\resizebox{\columnwidth}{!}{%
\begin{tabular}{lccc}\toprule
Model & Semantic$\uparrow$ & Patch$\uparrow$ & Motion ratio \\ \midrule
Self Forcing (zero-shot, T2V weights) & 0.510 & 0.399 & 4.57 \\
Teacher forcing & 0.717 & 0.613 & 1.70 \\
Dynamics Forcing & \textbf{0.804} & \textbf{0.733} & \textbf{1.00} \\ \bottomrule
\end{tabular}
}
\caption{Similarity between generated and real video on 50 held-out clips, first 5 seconds, scored with
DINOv2. Motion ratio: 1 = same amount of motion as the real video.}\label{tab:dyn}
\end{table}
\subsection{Qualitative results}
\begin{figure*}[t]\centering
\includegraphics[width=\linewidth]{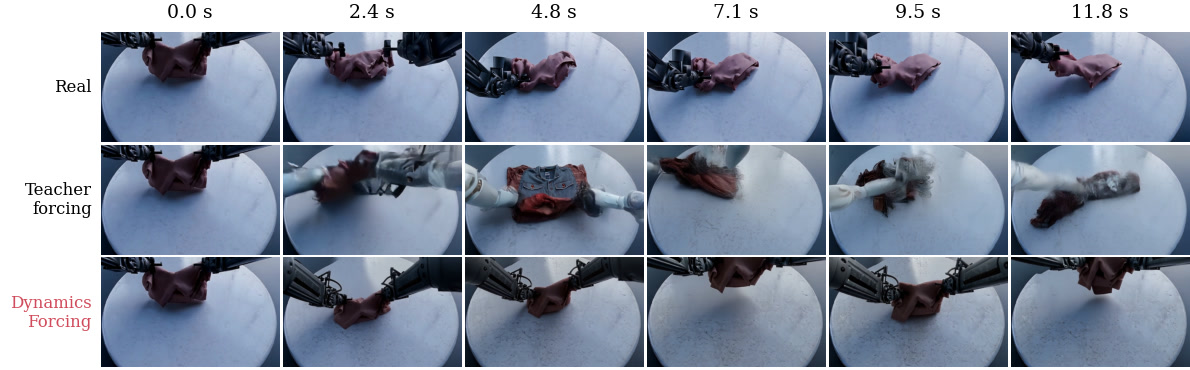}
\caption{Zero-shot held-out 12-second rollouts from the same first frame, instruction and noise. Neither
the evaluation trajectories nor these tasks appear in training. Top to bottom in each group: real video,
teacher forcing, Dynamics Forcing. The task is to fold the clothes.}\label{fig:qual}
\end{figure*}

\begin{figure*}[t]\centering
\includegraphics[width=\linewidth]{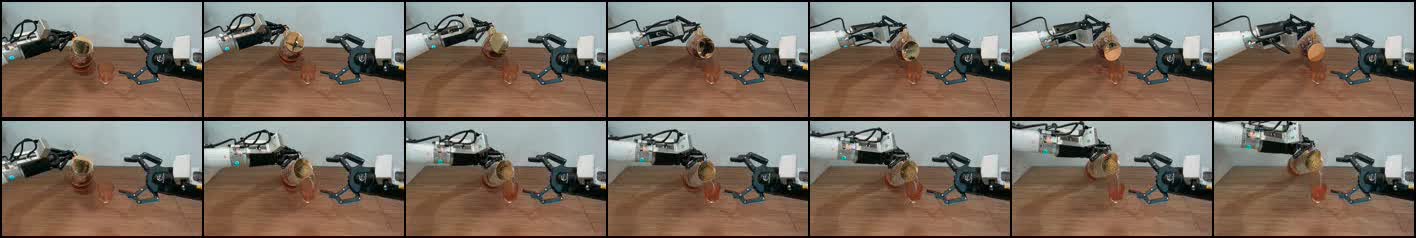}
\caption{Zero-shot pouring under the instruction ``pour water from the kettle into the glass.'' Top:
real continuation; bottom: Dynamics Forcing. Frames are sampled uniformly over the same interval. The
generated kettle tilts toward the glass, the water stream reaches it, and the liquid level rises as the
glass progressively fills, closely following the text instruction.}
\label{fig:pouring}
\end{figure*}

Figure~\ref{fig:qual} evaluates long-horizon appearance and interaction consistency, and
Figure~\ref{fig:pouring} evaluates whether generated motion produces the requested object-state change.
In the held-out cloth-folding rollout of Fig.~\ref{fig:qual}, the teacher-forced model replaces the robot in
the first frame with a different arm within the first seconds, and the scene then degrades: the cloth
turns into another garment and the inserted arm dissolves. Dynamics Forcing keeps the robot and the
objects of the first frame through the 12 seconds.

In the zero-shot rollout in Figure~\ref{fig:pouring}, the generated robot tilts the kettle, a continuous
stream reaches the glass, and the glass progressively fills as its liquid level rises. This interaction
is excluded from training, so the ordered action and state transition show that the model has learned
the causal change specified by the instruction rather than producing a static continuation or unrelated
motion. This result complements the aggregate motion and similarity scores in Table~\ref{tab:dyn}.

\subsection{Limitations}\label{sec:lim}
The current model is trained on 77 hours of video, a small data volume relative to the diversity and
precision required by long-horizon manipulation. Scaling training to a substantially larger public
video corpus is likely to improve the overall capability without changing the proposed objective.

Similarity still declines across the 12-second rollout, although Dynamics Forcing maintains higher
semantic similarity, patch similarity, and PSNR than teacher forcing in every evaluated segment.

Our current implementation supports rollouts of up to 15 seconds. Extending generation further requires
training and evaluating on longer sequences, which increases memory and computation; the present compute
budget limits this study to 15 seconds. Longer-context training is a direction for extending the
duration and stability of generation.

\section{Conclusion}
Dynamics Forcing trains a causal video world model from real video alone. Perturbing the history rather
than the target, and moving history and anchor guidance into training, reduces error accumulation and
discourages copying from recent context. On held-out manipulation videos, semantic and patch similarity
increase from 0.717 and 0.613 to 0.804 and 0.733, respectively. The same training improves the
VBench-I2V Total of the starting model from 85.15 to 90.02, indicating that the gains extend beyond the
robot-manipulation evaluation.

\FloatBarrier
\ifdefined\nolinenumbers\nolinenumbers\fi
{\small
\bibliographystyle{ieeenat_fullname}
\bibliography{refs}
}

\ifdefined\withsupplement
\twocolumn
\linenumbers
\appendix
\section{Implementation Details}\label{sec:implementation}
We train rank-64 LoRA adapters with scale 64 on the self-attention, cross-attention, and feed-forward
projections while keeping the pretrained backbone frozen. Optimization uses AdamW with learning rate
$10^{-4}$, $\beta_1=0$, $\beta_2=0.999$, zero weight decay, and a maximum gradient norm of 1.0. The
learning rate follows a linear decay to 10\% of its initial value. We use mixed precision and gradient
checkpointing with a per-device batch size of one.

The flow-matching noise schedule uses 1,000 diffusion timesteps with timestep shift 5. The VAE has a temporal stride
of 4, and each causal block contains eight latent frames; nine blocks therefore correspond to 12 seconds
at 24\,fps. At inference, classifier-free guidance is 5. The history-guidance branch is
applied to half of the training samples with weight 1 and history noise level $\sigma_h=0.5$; the
first-frame anchor branch uses $\sigma_a=0.8$. Self-resampled history uses three denoising steps and is
selected with probability 0.7. Joint denoising is selected with probability 0.5, uses a three-block
window, and samples the inter-block noise offset from $U(0.15,0.4)$ with noise levels clipped to
$[0.15,0.45]$.

\textbf{Data preprocessing.} Raw clips are resampled to 24\,fps, center-cropped to $832{\times}480$, and
encoded to 48-channel latents with the pretrained VAE before training; we train on latents and never
decode during the training loop. Text instructions are tokenized with the frozen text encoder of the
base model and dropped with probability 0.1 so the model does not rely on an always-present prompt.
For the held-out evaluations we keep the clips that contain a first-frame action (the robot or the hand
is already in contact or about to move), since those are the cases where predicting dynamics, rather
than predicting a static scene, is what is being measured.

\textbf{Training cost.} Training uses one 48\,GB GPU. The three history perturbations and the two
guidance branches add no learnable parameters. Their extra training cost comes from the gradient-free
weak-history prediction of Eq.~\eqref{eq:hg} and the three denoising steps used for self-resampling,
both of which run under {\tt torch.no\_grad}.

\textbf{Evaluation protocol.} All generation uses the same first frame, instruction, and initial noise
for every model in a comparison, and we score with the official VBench-I2V code and the frozen DINOv2
scorer described in the main text. No test clip, prompt, or VBench case is used for training, and we do
not sample multiple candidates and select the best; every number is from a single pass.

\section{Component Analysis}\label{sec:ablations}
We remove each component from the full Dynamics Forcing recipe and roll out the same held-out clips.
The observations below summarize the resulting failure modes.

\textbf{Random noised history.} Without it, the model regresses to copying the previous block: motion
ratio drops well below one and the arm freezes early. The schedule that ties history noise to the
target noise is important; noising history at a fixed, target-independent level instead makes the
history too clean when the target is noisy, and the copying shortcut reappears.

\textbf{Self-resampled history.} Removing resampling leaves only isotropic noise in the history, which
does not reproduce the correlated errors of autoregressive rollout; the model still drifts on long
sequences because it has never seen its own error pattern. Resampling alone, without random noising and
joint denoising, is also insufficient, because the model can copy its only mildly corrupted recent
blocks.

\textbf{Joint denoising.} Removing the three-block window makes the transition between consecutive
blocks abrupt and increases late-rollout drift. The lower bound on the inter-block noise keeps recent
history from being clean enough to copy, and the upper bound keeps it informative; raising the upper
bound from 0.45 to 0.6 made the robot arm fade, which is why we clip at $0.45$.

\textbf{Training-time guidance.} Removing history guidance restores appearance drift of objects over
long rollouts, and removing first-frame anchor guidance restores drift of the robot itself. Keeping
both branches inside training adds no guidance pass beyond the standard classifier-free-guidance
inference used by the starting model.

\section{Additional Qualitative Results}\label{sec:additional-qual}
Additional held-out examples complement Fig.~\ref{fig:qual}. They cover object displacement,
articulated motion, and long-horizon appearance preservation
(Figs.~\ref{fig:app-manip-a}--\ref{fig:app-manip-d}). For each example, all methods receive the same
first frame, instruction, and initial noise. Additional VBench-I2V examples compare Astronex-World and
Dynamics Forcing on the same official prompts. We include both high-motion and low-motion scenes to
show that the method suppresses unsupported motion without making all outputs static.

\begin{figure*}[t]\centering
\includegraphics[width=\linewidth]{figs/appendix_phys_12c995.jpg}
\caption{Held-out manipulation rollout with a visible object-state change. Rows show the real video,
teacher forcing, and Dynamics Forcing. Dynamics Forcing better preserves the robot and object geometry
while following the interaction. Frames are sampled uniformly in time.}\label{fig:app-phys-a}
\end{figure*}

\begin{figure*}[t]\centering
\includegraphics[width=\linewidth]{figs/appendix_phys_1963bd.jpg}
\caption{Additional held-out manipulation rollout. Rows show the real video, teacher forcing, and
Dynamics Forcing. The example tests articulated motion and the persistence of small objects over the
rollout. Frames are sampled uniformly in time.}\label{fig:app-phys-b}
\end{figure*}

\begin{figure*}[t]\centering
\includegraphics[width=\linewidth]{figs/appendix_manip_008a7d.jpg}
\caption{Held-out manipulation rollout with object displacement. Rows show the real video, teacher
forcing, and Dynamics Forcing. Teacher forcing loses the object identity and the scene degrades within
the first seconds, whereas Dynamics Forcing keeps the first-frame object and completes more of the
intended motion. Frames are sampled uniformly in time.}\label{fig:app-manip-a}
\end{figure*}

\begin{figure*}[t]\centering
\includegraphics[width=\linewidth]{figs/appendix_manip_01dc0b.jpg}
\caption{Held-out manipulation rollout with a long-horizon hand--object interaction. Rows show the real
video, teacher forcing, and Dynamics Forcing. This example stresses appearance preservation: the hand
and the manipulated object must stay recognisable across all nine blocks. Frames are sampled uniformly
in time.}\label{fig:app-manip-b}
\end{figure*}

\begin{figure*}[t]\centering
\includegraphics[width=\linewidth]{figs/appendix_manip_034655.jpg}
\caption{Held-out manipulation rollout with a gripper-based pick-and-place motion. Rows show the real
video, teacher forcing, and Dynamics Forcing. The gripper geometry is a failure mode for the
teacher-forced model and is preserved by Dynamics Forcing. Frames are sampled uniformly in
time.}\label{fig:app-manip-c}
\end{figure*}

\begin{figure*}[t]\centering
\includegraphics[width=\linewidth]{figs/appendix_manip_03df97.jpg}
\caption{Held-out manipulation rollout with articulated object motion. Rows show the real video,
teacher forcing, and Dynamics Forcing. Dynamics Forcing follows the object motion more faithfully while
keeping the robot and the background stable over the 12-second rollout. Frames are sampled uniformly in
time.}\label{fig:app-manip-d}
\end{figure*}

\begin{figure*}[t]\centering
\includegraphics[width=\linewidth]{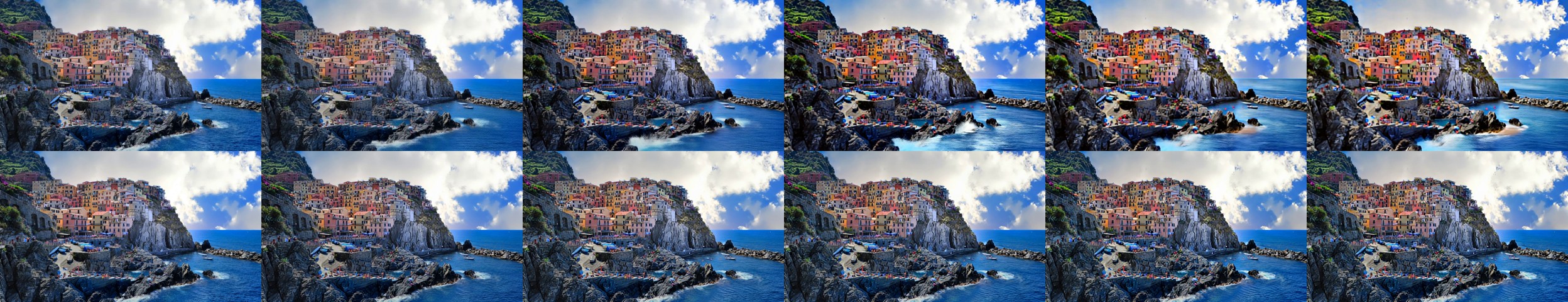}
\caption{Additional VBench-I2V comparison on ``colorful buildings on the seaside cliffs.'' Top:
Astronex-World; bottom: Dynamics Forcing. The base rollout remains stable in this example, allowing the
comparison to focus on fine-grained structure and temporal consistency rather than catastrophic drift.
Frames are sampled uniformly from each generated video.}\label{fig:app-vbench-cliffs}
\end{figure*}

\begin{figure*}[t]\centering
\includegraphics[width=\linewidth]{figs/appendix_vbench_lighthouse.jpg}
\caption{Additional VBench-I2V comparison on ``a large wave crashes into a lighthouse on a stormy
day.'' Top: Astronex-World; bottom: Dynamics Forcing. Frames are sampled uniformly from each generated
video.}\label{fig:app-vbench-lighthouse}
\end{figure*}

\begin{figure*}[t]\centering
\includegraphics[width=\linewidth]{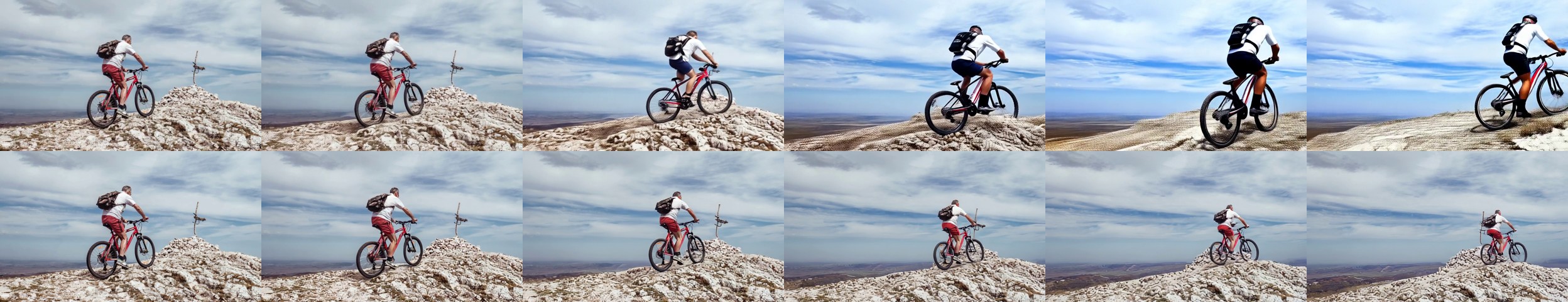}
\caption{Additional VBench-I2V comparison on ``a man riding a mountain bike on top of a rocky hill.''
Top: Astronex-World; bottom: Dynamics Forcing. Frames are sampled uniformly from each generated
video.}\label{fig:app-vbench-bike}
\end{figure*}

\begin{figure*}[t]\centering
\includegraphics[width=\linewidth]{figs/appendix_vbench_cargo_ship.jpg}
\caption{Additional VBench-I2V comparison on ``a large cargo ship sailing in the water.'' Top:
Astronex-World; bottom: Dynamics Forcing. Frames are sampled uniformly from each generated
video.}\label{fig:app-vbench-ship}
\end{figure*}

\begin{figure*}[t]\centering
\includegraphics[width=\linewidth]{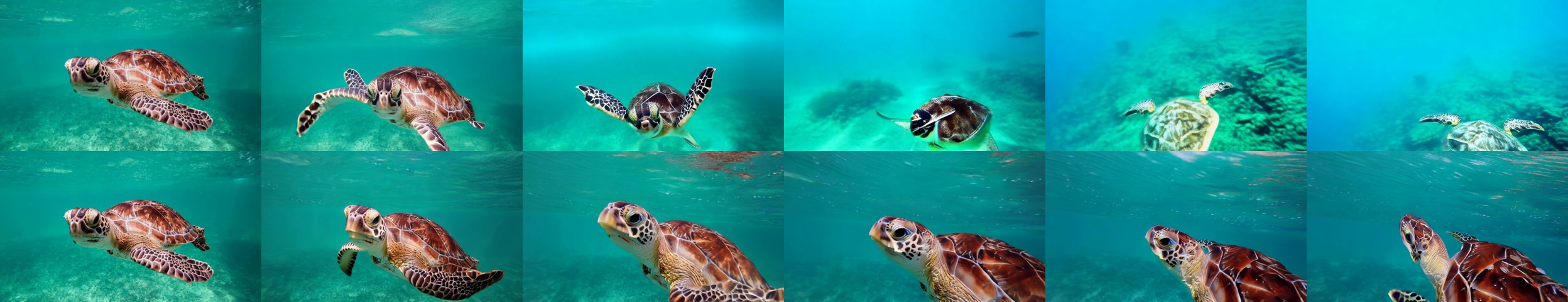}
\caption{Additional VBench-I2V comparison on ``a sea turtle swimming in the ocean under the water.''
Top: Astronex-World; bottom: Dynamics Forcing. Frames are sampled uniformly from each generated
video.}\label{fig:app-vbench-turtle}
\end{figure*}
\fi

\end{document}